\documentclass[11pt]{article}
\usepackage[utf8]{inputenc}
\usepackage{graphicx}
\usepackage{float}
\usepackage{animate}
\usepackage{comment,color,tikz}
\usepackage[hang,flushmargin]{footmisc}
\usetikzlibrary{matrix, positioning, arrows}
\tikzset{main node/.style={circle,fill=white,draw,minimum size=1cm,inner sep=0pt},}  
\usetikzlibrary{fit}
\tikzset{
  comp/.style = {
    minimum width  = 4cm,
    minimum height = 2.25cm,
    text width     = 4cm,
    inner sep      = 0pt,
    text           = purple,
    align          = center,
    font           = \large,
    transform shape,
    thick
  },
  monitor/.style = {draw = none, xscale = 18/16, yscale = 11/9},
  display/.style = {shading = axis, left color = black!60, right color = black},
  ut/.style      = {fill = gray}
}
\tikzset{
  computer/.pic = {
    \node(-m) [comp, pic actions, monitor]
      {\phantom{\parbox{\linewidth}{\tikzpictext}}};
    \node[comp, pic actions, display] {\tikzpictext};
    \begin{scope}[x = (-m.east), y = (-m.north)]
      \path[pic actions, draw = none]
        ([yshift=2\pgflinewidth]-0.1,-1) -- (-0.1,-1.3) -- (-1,-1.3) --
        (-1,-2.4) -- (1,-2.4) -- (1,-1.3) -- (0.1,-1.3) --
        ([yshift=2\pgflinewidth]0.1,-1);
      \path[ut]
        (-1,-2.4) rectangle (1,-1.3)
        (-0.9,-1.4) -- (-0.7,-2.3) -- (0.7,-2.3) -- (0.9,-1.4) -- cycle;
      \path[pic actions, fill = none]
        (-1,1) -- (-1,-1) -- (-0.1,-1) -- (-0.1,-1.3) -- (-1,-1.3) --
        (-1,-2.4) coordinate(sw)coordinate[pos=0.5] (-b west) --
        (1,-2.4) -- (1,-1.3) coordinate[pos=0.5] (-b east) --
        (0.1,-1.3) -- (0.1,-1) -- (1,-1) -- (1,1) -- cycle;
      \node(-c) [fit = (sw)(-m.north east), inner sep = 0pt] {};
    \end{scope}
  }
}
\usepackage{filecontents}

\usepackage{xcolor,amssymb}
\usepackage[backend=biber,style=authoryear]{biblatex}
 
\usepackage{amsmath}
\graphicspath{ {images/} }
\usepackage[top=1.2in, bottom=1.2in, left=1.2in, right=1.2in]{geometry}
\usepgflibrary{shapes.arrows}

\normalsize 

\title{\vspace{-2.0cm} Mathematical Decisions \& Non-causal Elements of Explainable AI}
\date{}

\author{Atoosa Kasirzadeh \\ \small{University of Toronto \& Australian National University} \\ atoosa.kasirzadeh@anu.edu.au}
\begin{document}
\maketitle


\vspace{1cm}

\noindent \textbf{\large Abstract.} The social implications of algorithmic decision-making in sensitive contexts have generated lively debates among multiple stakeholders, such as moral and political philosophers, computer scientists, and the public. Yet, the lack of a common language and a conceptual framework for an appropriate bridging of the moral, technical, and political aspects of the debate prevents the discussion to be as effective as it can be. Social scientists and psychologists are contributing to this debate by gathering a wealth of empirical data, yet a philosophical analysis of the social implications of algorithmic decision-making remains comparatively impoverished. In attempting to address this lacuna, this paper argues that a hierarchy of different types of explanations for why and how an algorithmic decision outcome is achieved can establish the relevant connection between the moral and technical aspects of algorithmic decision-making. In particular, I offer a multi-faceted conceptual framework for the explanations and the interpretations of algorithmic decisions, and I claim that this framework can lay the groundwork for a focused discussion among multiple stakeholders about the social implications of algorithmic decision-making, as well as AI governance and ethics more generally.

\newpage

\section{Introduction}

Governments and private actors are applying the most recent wave of AI, deep learning algorithms, in high-stake decisions.\footnote{This enterprise is motivated by the fact that context-sensitive decision-making can be thought to be an instance of performing algorithmic tasks such as pattern recognition, classification, or clustering.} These algorithms resolve several critical decision-making problems such as hiring employees, assigning loans and credit scores, making medical diagnoses, and dealing with criminal recidivism.\footfullcite{cabitza2017unintended,  angwin2016machine, bodo2017tackling, veale2018clarity} Hence, the deployed algorithms have begun to influence several aspects of our social lives. However, since these algorithms are opaque, one proposed approach towards legitimizing the incorporation of algorithms in several decision-making processes has been to require the algorithms to explain themselves.\footnote{The scope of AI in this paper is limited to deep supervised learning algorithms, due to the fact that the deployment of these algorithms in critical decision-making contexts has mainly motivated the recent concerns about the explainability and interpretability of AI. Although I find the discussions relevant, I do not directly address explainability and interpretability of logic-based symbolic manipulation paradigms, unsupervised, or non-supervised learning algorithms.}   

Building AI systems that are able to explain themselves (so-called ``explainable AI'') is motivated by several reasons: to increase the societal acceptance of the outcomes of algorithmic decisions, to establish trust in the results of these decisions (e.g., if one plans to take action based on an algorithm's prediction), to make algorithms accountable to the public, to validate these decisions, and to facilitate a fruitful conversation among different stakeholders concerning the justification of using these algorithms for decision-making. Therefore, the demands for AI systems that can explain their decisions is growing. These AI explanations are algorithmically generated, latch onto reality, and can be interpreted and understood by humans. But what is explainability and interpretability?

Computer scientists have suggested that AI explainability and interpretability are not monolithic concepts, and are used in different ways by these scientists.\footfullcite{lipton2016mythos,doshi2017towards} One promising approach for gaining conceptual clarification is to seek inspirations from philosophy about what an explanation is, and how an explanation relates to interpretation and understanding. This paper aims to achieve this goal. Thus far, the inspiration-seeking approach from philosophy has mainly focused on elucidating what the data-driven, causal aspects of an explanation are. For instance, in the most extensive survey in this area, Miller entirely dismisses the non-causal aspects of explanations:\footfullcite{miller2018explanation}

\begin{quote}
But what constitutes an explanation? This question has created a lot of debate in philosophy, but accounts of explanation both philosophical and psychology [sic] stress the importance of causality in explanation — that is, an explanation refers to causes [...]. There are, however, definitions of non-causal explanation [...]. These definitions [are] out of scope in this paper, and they present a different set of challenges to explainable AI. 
\end{quote}

In a similar vein, Mittelstadt limit the discussion of explainable AI to causal investigations:\footfullcite{mittelstadt2019explaining} 

\begin{quote}
Returning to philosophy, types of explanations can be distinguished according to their completeness, or the degree to which the entire causal chain and necessity of an event can be explained [...]. Often this is expressed as the difference between ‘scientific’ and ‘everyday’ explanations (both of which deal with causes of an event; e.g. Miller (2018), or ‘scientific’ (full) and ‘ordinary’ (partial) causal explanations [...].
\end{quote}

In this paper, I take an explanation to be a response to a why-question,\footfullcite{Bromberger1966} and to be somehow (empirically or mathematically) true. I set out to satisfy three objectives.

First, in Section 2, I argue that in the context of AI explanations, it is useful to distinguish between minimalist and maximalist conceptions of explanation. The former involves a variety of explanations, understandable by a specific group of people, about how black box AI make decisions.\footnote{In Section 4, I discuss the significance of the multiplicity of human's background assumptions to the goal of generating explanations.} The latter involves taking up insights from data analysis as well as the technical and mathematical processes by which an algorithm is governed, and through which the decision outcome is generated. While the minimalist view has been the main focus of the explainable AI discussion, I argue that we might ultimately move beyond minimalist conceptions of explanation if the aim of explainable AI is to increase the societal acceptance of algorithmic decisions, establish trust in the results of these decisions, and facilitate a fruitful conversation among different stakeholders on the justification of using these algorithms for decision-making. By incorporating some insights about non-data driven and mathematical aspects of explanation, rooted in the recent philosophical literature, I argue for the significance of mathematical, statistical, and optimality explanations.\footfullcite{potochnik2007optimality,bokulich2011scientific,batterman2014minimal,lange2016because,chirimuuta2017explanation,reutlinger2018explanation} I show how these explanations are helpful in addressing some important normative questions about the social use of AI decision-making. Hence, while I acknowledge the significance of data-driven information in some types of explanations, I argue that a comprehensive discussion about AI decision-making requires the identification of two additional explanations that are non-data driven, and are crucial to explaining why and how an algorithmic decision outcome is achieved: (1) the mathematical structures that underlie the representation of a decision-making situation, and (2) the statistical and optimality facts in terms of which the machine learning algorithm is designed and implemented. 



Second, in Section 3, I argue that a properly focused conversation about the societal, moral, and political impacts of algorithmic decision-making requires the acknowledgement of two pluralities: (i) the plurality of why-questions: there are different why-questions to ask about a particular decision outcome, and (ii) the plurality of explanatory responses: there are different responses to a particular why-question. Accordingly, I propose a more philosophically-informed explanatory schema in which a variety of data-driven and non-data driven information find their place in explaining AI decisions.\footnote{To capture the many differences and similarities pertaining to explainability and interpretability, I distinguish between the two issues. I take explanations to track some true facts about the world. Interpretations depend on the audiences' background assumptions for explanatory judgements. First, I treat these two issues separately (Sections 3 and 4), and then I bring them together (Section 5).} The explanatory schema is composed of different levels of explanation, and is sorted in ascending order of locality from the most non-data driven and structurally global to the most data-driven and local explanations.

Third, in Section 4, I discuss how background assumptions (such as goals and norms) held by the audiences of explanations can influence interpretations of the explanations. This suggests that the explanations for a given decision might be diverse in relation to different precedent assumptions of an audience. Accordingly, I propose an interpretation schema that stands in relation to the explanatory hierarchy. Each level of the interpretation schema incorporates a set of background assumptions held by the audience of an explanation that is required for understanding and interpreting an explanation. Hence, each level of the interpretation schema corresponds to a certain level of the explanatory hierarchy. The two explanatory hierarchy and interpretation schema lay out a conceptual framework for bringing a unified focus to the conversation among different stakeholders concerning the social implications of algorithmic decision-making. Such a conversation could include, but is not limited to, the critical assessment and analysis of the implicit moral, political, and technical assumptions embedded in the employment of an AI system in a particular sensitive decision context. The paper closes with some concluding remarks in Section 5.


\section{Mathematical decisions: structure, statistics, and optimization}

Let us consider a machine-learning algorithm that sifts through several job applications to recommend a future employee for company X. Nora, a competent candidate, applies for the job. Her application gets rejected by the algorithmic decision. Nora wants to know why she is rejected (Figure 1). She does not seek a just-so-story that somehow makes sense by organizing events into an intelligible whole. She wants the explanation to have factual foundations and to be somehow (empirically or mathematically) true. 

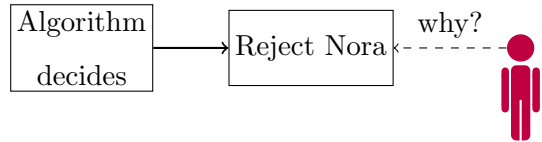
\begin{figure}[h]
\centering
\begin{tikzpicture}[every text node part/.style={align=center}]
    \node[main node, rectangle, inner sep=0.5ex] (1) {Algorithm \\ decides};
    \node[main node, rectangle, inner sep=0.5ex] (4) [right = 1cm of 1] {Reject Nora};
    \node[circle,fill=purple,minimum size=1mm] (head) [right=1.5cm of 4]{};
    \node[rounded corners=2pt,minimum height=1cm,minimum width=0.3mm,fill=purple,below = 1pt of head] (body) {};
    \draw[line width=1mm,purple,round cap-round cap] ([shift={(2pt,-1pt)}]body.north east) --++(-90:6mm);
    \draw[line width=1mm,purple,round cap-round cap] ([shift={(-2pt,-1pt)}]body.north west)--++(-90:6mm);
    \draw[thick,white,-round cap] (body.south) --++(90:5.5mm);
     \path[->,thick, below=8pt]
    (1) edge node {} (4);
    \path[->,dashed, right=8pt]
    (head) edge node [midway, above, sloped]{why?} (4); 
\end{tikzpicture}

\caption{Algorithm decides and a why-question.}
\end{figure}

This decision-making problem is an exemplar of the social difficulties that might be resolved using a machine-learning algorithm. Although there are many different kinds of learning algorithms, such as supervised, unsupervised, and reinforcement algorithms. In what follows, I will only focus on explanations in the context of deep supervised learning. This assumption is also justified on the ground that several critical decision-making problems are instances of classification or regression, two tasks performed by supervised learning algorithms.\footnote{It should be noted that these insights with respect to classification and regression can be generalized and extended to other learning algorithms.} A brief review of the basics of deep supervised learning and the networks on which these algorithms learn facilitates the rest of our discussion. Through the statistical and optimality analysis of a massive corpus of data, without receiving explicit rules, supervised learning algorithms are able to learn about some structures and interrelations within the data. Using this learned information, they make predictions or determinations about new data. A supervised learning algorithm is trained on a large data set of input-output pairs. The output elements are taken to be the right answers generated by an unknown function $g(x)$. A supervised algorithm seeks to find a hypothesis function $f(x)$ that approximates the alleged true function $g(x)$. The algorithm learns and generates $f(x)$ by estimating new parameters for several interactions among input features. This estimation is usually cached out in terms of the (total) minimization of some error as the difference among the data points of a given corpus and $f(x)$. The algorithm then uses $f(x)$ to predict and recommend outputs for new input data. Figure 2 illustrates a more fine-grained characterization of Figure 1, where a supervised learning algorithm is situated in a decision context and the why-question about the implemented algorithmic outcome arises. 

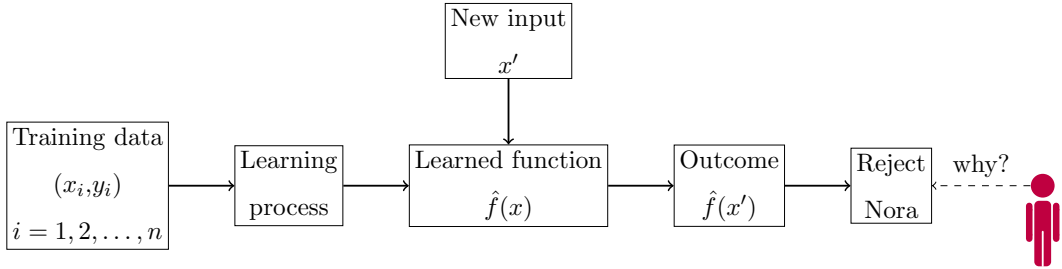
\begin{figure}[h]
\centering
\resizebox{14cm}{!}{\begin{tikzpicture}[every text node part/.style={align=center}]
    \node[main node, rectangle, inner sep=0.5ex] (1) {Training data \\ ($x_i$,$y_i$) \\$i=1,2,\dots,n$};
    \node[main node, rectangle, inner sep=0.5ex] (2) [right = 1cm of 1] {Learning \\ process};
    \node[main node, rectangle, inner sep=0.5ex] (3) [right = 1cm of 2]  {Learned function \\ $\hat{f}(x)$};
    \node[main node, rectangle, inner sep=0.5ex] (4) [right = 1cm of 3] {Outcome \\ $\hat{f}(x^{\prime}$)};
    \node[main node, rectangle, inner sep=0.5ex] (5) [above = 1cm of 3]  {New input \\ $x^{\prime}$};
    \node[main node, rectangle, inner sep=0.5ex] (6) [right = 1cm of 4]  {Reject \\ Nora};
    \node[circle,fill=purple,minimum size=1mm] (head) [right=1.5cm of 6]{};
    \node[rounded corners=2pt,minimum height=1cm,minimum width=0.3mm,fill=purple,below = 1pt of head] (body) {};
    \draw[line width=1mm,purple,round cap-round cap] ([shift={(2pt,-1pt)}]body.north east) --++(-90:6mm);
    \draw[line width=1mm,purple,round cap-round cap] ([shift={(-2pt,-1pt)}]body.north west)--++(-90:6mm);
    \draw[thick,white,-round cap] (body.south) --++(90:5.5mm);
    \path[->,thick, above=8pt]
    (1) edge node {} (2);
    \path[->,thick, right=8pt]
    (2) edge node {} (3);  
     \path[->,thick, below=8pt]
    (3) edge node {} (4);
    \path[->,dashed, right=8pt]
    (head) edge node [midway, above, sloped]{why?} (6); 
     \path[->,thick, right=8pt]
    (5) edge node {} (3);
     \path[->,thick, right=8pt]
    (4) edge node {} (6);
\end{tikzpicture}}
\caption{Supervised machine learning and a why-question.}
\end{figure}

This sketch illustrates the algorithmic elements that give rise to a decision outcome. Related to Nora's hiring example, the algorithm is designed to optimize some success criteria, such as minimizing cost, for a \emph{chosen} target variable. Then, if employee turnover is costly, a good employee might be defined as one who is likely to stay at the company for a long time. In other words, the target variable is longevity. 

We can now list a set of five explanatory questions that provide a more fine-grained questionnaire for why Nora is rejected. Responding to each of these questions provides information for why Nora is rejected: (1) What is a causal explanation, rooted in data analysis, of Nora's rejection? That is, what input features, representing Nora, caused Nora's rejection? (2) What is a correlational explanation, rooted in data analysis, for Nora's rejection? That is, what input features were correlated with the decision outcome for Nora's rejection? (3) What is a statistical and optimization-based explanation for rejecting Nora? That is, what statistical assumptions and optimality functions during training contributed to determining the decision-making about Nora's case? (4) What role does the specific training data set play in Nora's rejection? (5) What is the set of Nora's quantifiable features that contribute to the algorithmic hiring decision? The plurality of the relevant explanatory questions, therefore, expands, as Figure 3 illustrates. Responses to the questions (1)--(5) explains maximally, from an algorithmic point of view, how and why the particular decision about Nora is obtained.

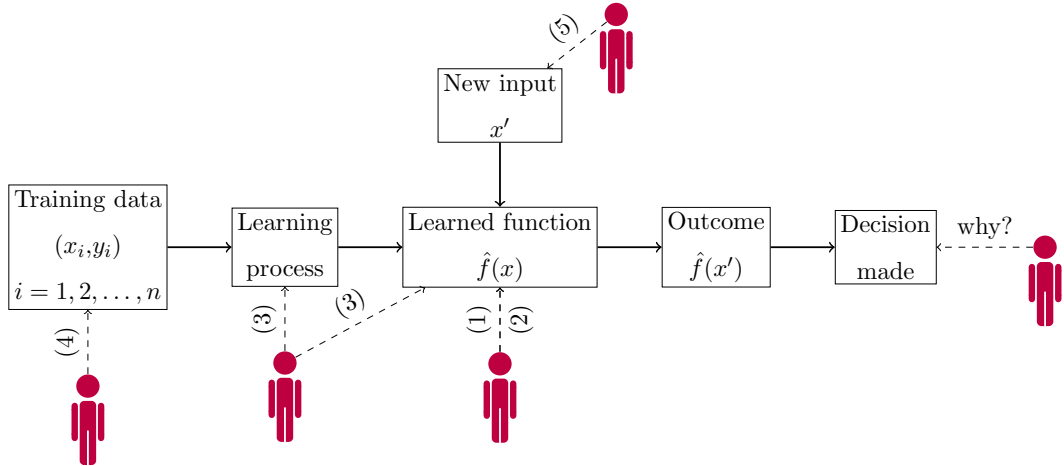
\begin{figure}[h]
\centering
\resizebox{14cm}{!}{\begin{tikzpicture}[every text node part/.style={align=center}]
    \node[main node, rectangle, inner sep=0.5ex] (1) {Training data \\ ($x_i$,$y_i$) \\$i=1,2,\dots,n$};
    \node[main node, rectangle, inner sep=0.5ex] (2) [right = 1cm of 1] {Learning \\ process};
    \node[main node, rectangle, inner sep=0.5ex] (3) [right = 1cm of 2]  {Learned function \\ $\hat{f}(x)$};
    \node[main node, rectangle, inner sep=0.5ex] (4) [right = 1cm of 3] {Outcome \\ $\hat{f}(x^\prime)$};
    \node[main node, rectangle, inner sep=0.5ex] (5) [right = 1cm of 4] {Decision \\ made};
    \node[circle,fill=purple,minimum size=1mm] (head) [right=1.5cm of 5]{};
    \node[rounded corners=2pt,minimum height=1cm,minimum width=0.3mm,fill=purple,below = 1pt of head] (body) {};
    \draw[line width=1mm,purple,round cap-round cap] ([shift={(2pt,-1pt)}]body.north east) --++(-90:6mm);
    \draw[line width=1mm,purple,round cap-round cap] ([shift={(-2pt,-1pt)}]body.north west)--++(-90:6mm);
    \draw[thick,white,-round cap] (body.south) --++(90:5.5mm);
    
    \node[circle,fill=purple,minimum size=1mm] (head2) [below = 1cm of 1]{};
    \node[rounded corners=2pt,minimum height=1cm,minimum width=0.5mm,fill=purple,below = 1pt of head2] (body2) {};
    \draw[line width=1mm,purple,round cap-round cap] ([shift={(2pt,-1pt)}]body2.north east) --++(-90:6mm);
    \draw[line width=1mm,purple,round cap-round cap] ([shift={(-2pt,-1pt)}]body2.north west)--++(-90:6mm);
    \draw[thick,white,-round cap] (body2.south) --++(90:5.5mm);
    
    \node[circle,fill=purple,minimum size=1mm] (head3) [below = 1cm of 2]{};
    \node[rounded corners=2pt,minimum height=1cm,minimum width=0.5mm,fill=purple,below = 1pt of head3] (body3) {};
    \draw[line width=1mm,purple,round cap-round cap] ([shift={(2pt,-1pt)}]body3.north east) --++(-90:6mm);
    \draw[line width=1mm,purple,round cap-round cap] ([shift={(-2pt,-1pt)}]body3.north west)--++(-90:6mm);
    \draw[thick,white,-round cap] (body3.south) --++(90:5.5mm);
    
    \node[circle,fill=purple,minimum size=1mm] (head4) [below = 1cm of 3]{};
    \node[rounded corners=2pt,minimum height=1cm,minimum width=0.5mm,fill=purple,below = 1pt of head4] (body4) {};
    \draw[line width=1mm,purple,round cap-round cap] ([shift={(2pt,-1pt)}]body4.north east) --++(-90:6mm);
    \draw[line width=1mm,purple,round cap-round cap] ([shift={(-2pt,-1pt)}]body4.north west)--++(-90:6mm);
    \draw[thick,white,-round cap] (body4.south) --++(90:5.5mm);
   
    \node[main node, rectangle, inner sep=0.5ex] (6) [above = 1cm of 3]  {New input \\ $x^\prime$}; 
    
    \node[circle,fill=purple,minimum size=1mm] (head5) [above right = 1cm of 6]{};
    \node[rounded corners=2pt,minimum height=1cm,minimum width=0.5mm,fill=purple,below = 1pt of head5] (body5) {};
    \draw[line width=1mm,purple,round cap-round cap] ([shift={(2pt,-1pt)}]body5.north east) --++(-90:6mm);
    \draw[line width=1mm,purple,round cap-round cap] ([shift={(-2pt,-1pt)}]body5.north west)--++(-90:6mm);
    \draw[thick,white,-round cap] (body5.south) --++(90:5.5mm);

    \path[->,thick, above=8pt]
    (1) edge node {} (2);
    \path[->,thick, right=8pt]
    (2) edge node {} (3);  
     \path[->,thick, below=8pt]
    (3) edge node {} (4);
    \path[->,thick, below=8pt]
    (4) edge node {} (5);
    \path[->,dashed, right=8pt]
    (head) edge node [midway, above, sloped]{why?} (5); 
     \path[->,thick, right=8pt]
    (6) edge node {} (3);
    
    \path[->,dashed, right=8pt]
    (head2) edge node [midway, above, sloped]{(4)} (1); 
    
    \path[->,dashed, right=8pt]
    (head3) edge node [midway, above, sloped]{(3)} (2);
    
    \path[->,dashed, right=8pt]
    (head3) edge node [midway, above, sloped]{(3)} (3);
    
    \path[->,dashed, right=8pt]
    (head4) edge node [midway, above, sloped]{(1)} (3); 
    
    \path[->,dashed, right=8pt]
    (head4) edge node [midway, below, sloped]{(2)} (3); 
    
    \path[->,dashed, right=8pt]
    (head5) edge node [midway, above, sloped]{(5)} (6); 

\end{tikzpicture}}

\caption{The plurality of why-questions about an algorithmic decision.}
\end{figure}


Deep supervised learning algorithms usually demonstrate their learning through a high-dimensional and highly nonlinear function that is incomprehensible for humans. For this reason, several methods (e.g., models) have been devised to explain why and how these algorithms generate a particular decision outcome. Thus far, the work on the explainability of deep supervised learning has largely focused on constructing methods that answer questions of types (1) or (2): what input features caused Nora's rejection? What input features correlated with the decision outcome for Nora's rejection? These methods reveal that there is a kind of information rooted in data analysis, whether it is causal, correlational, or contrastive, about the relationship among the salient features that contribute to the algorithm's decision outcome. Roughly, the landscape of research on making AI explainable and interpretable has been divided into three groups (i)--(iii) based on the kind of explanation that is generated by these methods. 

(i) Correlational explanations reveal some approximation for the relationship among data points or specific features. These relations might be revealed using a particular model, such as a linear regression. A linear model approximates how the feature of an instance (e.g., Nora's longevity at a work place, or the temperature) is associated with the predicted outcome (e.g., Nora's duration of stay at a job, the number of bikers in the streets). Contrariwise, the correlational explanations might be obtained in a model-agnostic and local way by tweaking the input features and tracing their contributions to output features.\footfullcite{ribeiro2016should} For instance, without using a specific model approximating the workings of a deep learning algorithm, a framework for local data analysis can reveal how Nora's gender and her previous longevity at a work place correlates with her hire. A correlational explanation reveals information, as the name suggests, about the correlation among input features and the outcome of a decision. (ii) Causal explanations reveal some causal relations among data-points via some sort of causal modelling.\footfullcite{zhao2019causal, lakkaraju2017learning}. For instance, a causal explanation for why Nora was rejected might be that some of her features, such as her ethnicity, have caused her rejection. (iii) Example-based explanations, such as counterfactual and prototype explanations, compare and contrast a specific predicted output by finding similar outputs while changing some of the input features for which the predicted outcome changes in a relevant way.\footfullcite{kim2016examples, wachter2017counterfactual} These explanations provide contrastive information rooted in data analysis, that is, by providing counterfactual statements about what would have been needed to be different about an individual’s situation to get a different, preferred outcome. For instance, Aron, who had a very similar CV to Nora, but who had a longer job longevity, was granted the job.

Explanations of kind (i)--(iii) reveal causal and correlational information in relation to the quantified features of an individual. They give responses to the explanatory questions (1) and (2). However, the responses to these two questions do not fully reveal why this algorithmic decision outcome is obtained. Merely responding to the questions (1) and (2) do not satisfy the multiple purposes that the researchers on the explainable and interpretable AI aim to achieve: to increase societal acceptance of algorithmic decision outcomes, to establish trust in the results of these decisions, to generate human-level transparency about why a decision outcome is achieved, and to have a fruitful conversation among different stakeholders concerning the justification of using these algorithms for decision-making.

In the rest of this section, I argue for the significance of two other kinds of explanations, in addition to the correlational and causal explanations discussed above. These two kinds of explanations are framed in terms of mathematical and statistical-optimality rules, which govern the algorithmic decision procedure and warrant the decision outcome. These explanations reveal how a decision is achieved due to the structural, statistical, and optimality facts that constitute the algorithmic process. They give responses to explanatory questions (3)--(5), and they cast light on the motivation for a maximalist conception of AI explainability. I begin by defending the importance of structural explanations for algorithmic decision-making.



\subsection{Representing on mathematical structures}

The producer of an explanation for a given decision stands in a relation with the decision-making problem. This problem is represented in a particular way using a set of concepts. Hence, an explanation for a decision outcome is generated employing the elements and concepts by which a decision-making problem is represented. The generated explanation for why a decision outcome is achieved is the result of an investigation and a reflection on the reasons for why an algorithmic decision outcome occurred. As the decisions of interest to this paper are fundamentally represented on the deep neural network structure, I first say a few words about the mathematical structure of deep neural networks.

In a successful incarnation, deep learning algorithms represent the problem according to the framework of artificial neural networks.\footfullcite{lecun2015deep} These networks are inspired by the topological structure and function of the human brain's architecture. Artificial neural networks are composed of a set of connected nodes (neurons) and a set of arcs that connect the nodes and transfer signals among them. These nodes and arcs are organized into multi-layers. Typically, an artificial neural network has an input layer, an output layer, and multiple layers between the input and output layers. An artificial neural network is deep if there are many hidden layers between input and output layers. Each node of the input layer represents an individual feature from each sample of the training data set. For example, Nora's length of employment at her previous jobs, her age, and her education level can be part of a set of input features. A numerical weight is associated with each arc which represents the strength of the connection between the nodes. Corresponding to each node of the deep and output layers, there is an activation function that receives the sum of the weights into that node and outputs a value into a neuron in the next layer. Learning occurs when an optimization method such as gradient descent statistically minimizes an error between the algorithmic outcome and a true output. Figure 4 illustrates a simplistic sketch of this architecture, with five input features and two hidden layers. 

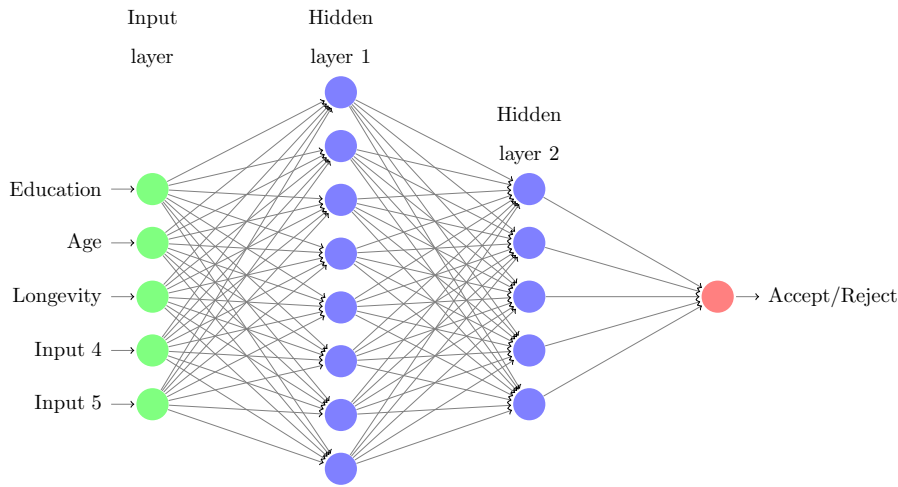
\begin{figure}
\def\layersep{3.5cm}
\centering
\resizebox{12cm}{!}{\begin{tikzpicture}[
   shorten >=1pt,->,
   draw=black!50,
    node distance=\layersep,
    every pin edge/.style={<-,shorten <=1pt},
    neuron/.style={circle,fill=black!25,minimum size=17pt,inner sep=0pt},
    input neuron/.style={neuron, fill=green!50},
    output neuron/.style={neuron, fill=red!50},
    hidden neuron/.style={neuron, fill=blue!50},
    annot/.style={text width=4em, text centered}
]

    \foreach \name  [count=\y] in 
    {Education, Age, Longevity, Input 4, Input 5}
     {   \node[input neuron, pin=left:\name] (I-\y) at (0,-\y cm) {};  
     }

    \newcommand\Nhidden{2}
    \newcommand\NodOne{8}
    \newcommand\NodTwo{5}
    \newcommand\Nod{5}


     \foreach \y in {1,...,\NodOne} {
          \path[yshift=1.80cm]
              node[hidden neuron] (H1-\y) at (1*\layersep,-\y cm) {};
              }
    \node[annot,above of=H1-1, node distance=1cm] (hl1) {Hidden layer 1};
     \foreach \y in {1,...,\NodTwo} {
          \path[yshift=0cm]
              node[hidden neuron] (H2-\y) at (2*\layersep,-\y cm) {};            
           }
    \node[annot,above of=H2-1, node distance=1cm] (hl2) {Hidden layer 2};          



    \node[output neuron,pin={[pin edge={->}]right:Accept/Reject}, right of=H\Nhidden-3] (O) {};

    \foreach \source in {1,...,5}{
        \foreach \dest in {1,...,\NodOne}{
            \path (I-\source) edge (H1-\dest);
         }
    }
    \foreach [remember=\N as \lastN (initially 1)] \N in {2,...,\Nhidden}
       \foreach \source in {1,...,\NodOne}
           \foreach \dest in {1,...,\NodTwo}
               \path (H\lastN-\source) edge (H\N-\dest);

    \foreach \source in {1,...,\NodTwo}
        \path (H\Nhidden-\source) edge (O);


    \node[annot,left of=hl1] {Input layer};
\end{tikzpicture}}
\caption{A simple sketch of the deep neural network architecture.}
\end{figure}

The deep learning architecture makes an implicit assumption for representing the decision problem: there is a morphism between the input layer of the mathematical representation and the source of the input layer data coming from the real world that makes up the significant elements of the decision-making situation. Using deep neural networks for decision-making requires that the significant elements of the feature space (e.g., Nora's features), relevant to the decision-making situation, should be quantifiable and measurable. To represent something mathematically is to have clear and distinct ideas about these elements and also the relation between them: what features are important and relevant? Moreover, instantiating these algorithms requires making several pragmatic decisions that guarantees the algorithmic functionality, for instance, how the decision-making configurations should be translated and reduced into numerical values on the neural network structure (e.g., the link between the nodes and what the activation function should be).


Mathematical structure and the form of representation as required by the algorithm gives a response to the explanatory question (5): What is the set of Nora’s quantifiable features that contribute to the algorithmic hiring decision? This answer reveals the following: the precondition for using these algorithms as decision-making tools is to reduce the relevant features of Nora, as a person, to quantifiable variables, measure them, and then feed them as the input data to the algorithms. This explanatory answer, as I will argue in details in Section 3, is tied to the response to the following morally-significant question concerning the \emph{acceptability} of this explanation: why does the algorithm merely observe Nora's attributes mathematically relevant to the decision-making context?

If the representation of the decision-making situation would be very different (e.g., represent the decision problem by non-measurable features), we might get a different decision outcome. Thus far, I have shown that a maximalist view of AI explanations admits that a part of the explanations for the algorithmic decision outcomes must be framed in terms of the mathematical structures that characterize the decision-making situation. Having established the explanatory relevance of the mathematical representation of the decision-making situation to algorithmic decision-making, I will now turn to the explanatory role of statistics and optimization in warranting AI decision outcomes.


\subsection{Statistics and optimization}

Deep supervised learning algorithms employ some amalgamation of statistical analysis and optimization in order to predict the probability of the occurrence of an outcome. First, I begin by a simple example in order to fix the relevant intuitions about the explanatory power of statistical rules, e.g., the case of pricing in the insurance industry. 

Statistical methods are tailored for calculation of risk premiums based on the past behaviour of a population, rather than a single individual. The cost of one's insurance policy depends on one's risk. This risk reflects the likelihood of making an insurance claim. The lower one's risk is, the lower one's premium will be. Insurers heavily rely on statistical rules about the law of large numbers and the related central limit theorem in order to decide about how much they may have to pay out by calculating the risk premiums.\footnote{The law of large numbers holds that the average of a large number of independent identically distributed random variables tends to fall close to the expected value. The central limit theorem, roughly, can be stated as follows: if X, a random variable, is the sum of a large number of independent and identically distributed random variables, then no matter how the identically distributed variables are distributed, X would have a Gaussian distribution.} For instance, the central limit theorem warrants using the Gaussian distribution for the estimation and likelihood definition when we consider the sums of independent and identically distributed random variables. Therefore, the decision outcomes of the deep supervised learning algorithms are partially dependent on statistical rules that are constitutive of learning from data sets. The law of large numbers and the central limit theorem, as two statistical rules, are used to explain the pooling of losses as an insurance mechanism, and they explain why an increase in the number of policyholders strengthens the sustainability of insurance policies by reducing the probability that the pool will fail.\footfullcite{smith1994law}

Similarly, concerning a supervised learning algorithm such as classification, the explanation for why a tumor is classified as malignant, rather than as benign, is determined by highly probable similarities to an analyzed statistical sample that suggest the tumor is like the other malignant tumors. These probabilistic claims are warranted due to accepting probability theory and statistical rules such as the law of large numbers and the central limit theorem as legitimate tools in drawing an inference. These statistical rules partially govern and influence the design of the decision procedure, and they therefore are an explanatory element for why a decision outcome, due to a statistical inference, is achieved. Back to Nora's rejection, the outcome recommendation is obtained based on the warranted statistical path (e.g., a specific Gaussian distribution). Having briefly established the explanatory relevance of statistical facts in relation to an algorithmic decision outcome, I now move on to make a case for the constitutive explanatory role of mathematical optimization for machine learning decisions.


Most supervised learning algorithms are based on optimizing a particular objective function. Let us discuss in details how mathematics becomes constitutive of the decision-making situation in the optimization step. First, mathematics is used in a variety of ways during the training of an artificial neural network. To train an artificial neural network is to resolve an optimization problem. For supervised learning algorithms, resolving an optimization problem on the deep neural networks often means having to calculate the weights on the arcs of the network such that a total loss function defined for the network is minimized. The loss function is a mathematical function, such as mean squared error $\frac{1}{n}\sum\limits_{i=1}^{n} (y_i-\hat{f}(x_i))^2$, that minimizes the difference between what the computed output should be and what the model has predicted. The minimization of a loss function happens by continuously updating the weights. The total loss of the network is the sum of the loss over all output layer nodes, or the composition of all loss functions on the nodes of the output layer. The most frequently used optimizer for minimizing the loss function in artificial neural networks is the stochastic gradient descent. This method updates the weights on the arcs of the network by calculating the gradient (derivative) of the loss function for each weight. The calculation is done by back propagation.\footfullcite{rumelhart1988learning}

Back propagation itself is based on four fundamental mathematical equations.\footfullcite{Nielsen2015} Why should we trust these equations rather than other equations? Because we can mathematically prove that the equations are true. In other words, we can prove that these equations are all consequences of the chain rule from multivariable calculus. Hence, mathematical proofs guarantee the efficiency of the use of back propagation for a decision-making problem. This suggests that if some optimality facts warrant that an algorithm can mathematically arrive at a decision, they are required to be used in an (maximalist) explanation for why this decision outcome is obtained.

In what follows, I briefly discuss two potential limits of statistical and optimality thinking styles for making decisions. These limits show that if a different non-statistical style of thinking is used for a decision-making situation, the decision outcome might be different.

First, humans are more efficient in learning abstractions through explicit, verbal definitions.\footfullcite{marcus2018deep} Deep learning lacks such a learning capability. Therefore, deep supervised decision-making is different from human decision-making, if humans base their decision-making on thinking styles that are based on learning through explicit, verbal definitions. Second, it is not straightforward how deep learning algorithms can incorporate prior knowledge into their reasoning style. As mentioned earlier, deep algorithms typically learn on a training database, in which there are sets of inputs associated with respective outputs. These algorithms learn all that is required for the problem by learning the relations between those inputs and outputs, using whatever clever architectural variants one might devise, along with techniques for cleaning and augmenting the data set. The incorporation of prior knowledge into the functioning of these algorithms is often missed. While there are decision-making contexts in which prior knowledge is significant to decision-making by humans, deep learning algorithms appear incapable of making decisions based on such human reasoning styles. This point will be clarified in an example shortly.

Of particular importance, too, is the role of human judgement in arriving at a particular solution to a decision-making problem. Smith distinguishes between two kinds of discernment capacities that are different between humans and AI systems.\footfullcite{Cantwellsmith} Smith reserves the term ``reckoning'' for the calculative rationality of present-day algorithms. This capability is empty of any ethical commitment, authenticity, or deep contextual awareness. On the other hand, the term ``judgement'' refers to the human capacity to understand the relations between the appearances and reality in an authentic way that grounds one's contextual awareness. This authentic understanding of the decision-making situation, which an AI lacks, rather than a calculative mathematical optimization might directly impact how humans might make a decision, and it might impact the outcome of a human decision-making process. In what follows, I give a real-life simple example of uniquely human judgemental capacity that I think highlights how such unique judgemental capacities, due to its authentic features, cannot be taught to a deep supervised neural network.

In April, 2019, a 96-year-old man, Mr. Coella, was charged with exceeding the speed limit in a school zone in Rhode Island.\footfullcite{coella2019} In court, Mr. Coella explained that he is taking care of his 63-year-old, handicapped son who has got cancer, and the reason for exceeding the speed limit is that he had been driving his son to the doctor's office when his car's speed exceeded. The judge, rather than appealing to the laws for ticket-giving to Mr. Coella's offense case, decided that Mr. Coella is ``a good man'', and that Mr. Coella's actions are ``what America is all about''. Accordingly, the judge dismissed Mr. Coella's ticket. It is extremely difficult, and perhaps impossible, to assume that the values captured by the ``goodness of a man'' and ``what America is all about'' can be meaningfully quantified and represented on the deep neural network architecture.


In summary, I have discussed that in addition to data-driven explanations for AI decision outcomes, there are also non-data driven and non-causal elements of mathematical structure, statistics, and mathematical optimization that directly influence the outcome of an AI decision-making problem, and hence explain why a decision outcome is achieved. Due to the fact that these elements make up the decision-making outcome, they must also be manifest in the explanations for why a decision is made. Let us say more about how the characterization of these non-data driven and mathematical explanations can be grounded by looking at the recent philosophical literature.

\subsection{Mathematical explanations}

The question of ``what is an explanation?'' has taken center stage in contemporary philosophy of science. In particular, philosophers of science have been extensively concerned with the analysis of explanation ever since Hempel and Oppenheim.\footfullcite{hempel1948studies} Roughly, these discussions aim to answer one of the following three questions. (1) Are explanations reducible to causal explanations, or are there genuine cases of non-causal explanations? (2) Whether, and if so how, can we specify the necessary and sufficient conditions for explanations? (3) How do non-causal explanations work, if they exist at all?

Hempel and Oppenheim offer an account of explanation that emphasizes its argumentative nature.\footfullcite{hempel1948studies, hempel1965aspects} According to that account, explanations have two constituent parts: the thing to be explained, or explanandum, and the explaining thing, or explanans. Hempel and Oppenheim stipulate that the explanandum must logically follow from the explanans which contains at least one law-like generalization. The argumentative nature of explanations generated by AI systems has been emphasized in the literature by Miller and Mittelstadt, among others.\footfullcite{miller2018explanation, mittelstadt2019explaining} Hempel's account, however, was too permissive and it allowed for irrelevant generalizations to be counted as explanations. 

Since Hempel, some philosophers have argued that a missing element of an account of explanation is causal information. Among many, Salmon and Strevens hold that explanations provide information about causal relations that are out there in the world.\footfullcite{salmon1984scientific, strevens2008depth} But this has not been the end of the story about what information should be included within an explanation.


More recently, the discussion about what explanations are has tilted towards unveiling the non-causal elements engaged in the production of an explanation such as the explanatory roles of mathematics.\footfullcite{batterman2001devil,lange2016because,chirimuuta2017explanation} A simple example reveals the insights behind these discussions. To explain why a mother cannot divide 23 strawberries among her three children, one can appeal to the mathematical fact that 23 cannot be divided 3 evenly. Of particular interest are cases of optimality explanations in which reference to an optimality notion, such as equilibrium, is responsible for an explanation of some empirical phenomena such as natural selection.\footfullcite{potochnik2007optimality,rice2015moving} Philosophical views about mathematical explanations extend to the explanation in decision-making sciences as well. Here is an example.

If a linear programming algorithm is used for a nurse scheduling, the reason for why a particular set of nurse schedules is obtained is partly due to the use of mathematical optimization for solving a linear programming problem such as the simplex method for arriving at the schedules. In a similar vein, as optimization algorithms are situated at the heart of machine learning, the producer of an explanation for decision-making needs to refer to the optimization as an important element in shaping the procedure and the outcome of decision-making. By elucidating the varieties of non-data driven and mathematical explanations, I have now sufficient resources to propose a hierarchy of different types of explanations for AI decision-making. 


\section{The many faces of an AI explanation}

The hierarchy of AI explanations, as illustrated in Figure 5, is composed of multiple levels. On the top level, there is a structural explanation, which reveals why a particular decision output is generated in virtue of a specific structural mapping of the salient features of a decision-making situation on to the input layer of the neural network. On a lower level, there is the statistical and optimality explanation that emphasizes the statistical and optimality laws or facts engaged in learning and forming the decision-making procedure. These two kinds of explanations together acknowledge the importance of non-data driven and mathematical elements that are constitutive and the ground for warranting AI decision outcomes. As we will discuss shortly, these explanations open a space for asking questions concerning the social, moral, and political assumptions for algorithmic decision-making. For example, we can ask about the reductive representation of significant decision-making attributes and the amalgamation of statistical and optimality thinking for critical decision-making problems affecting particular individuals. It is correct that these explanations might not be understandable by a lay-person. However, this reasoning should not make us dismissive of the significance of these explanations as to why an algorithmic decision outcome has been achieved. As I elaborate in Section 4, the importance of understanding the social, moral, and political implications of AI decision-making prevents us from setting minimal requirements on overall generation of explanations understandable by lay people, here and now. 

\tikzstyle{my arrow} = [draw=cyan!75, very thick, single arrow, minimum height=5.5cm, shape border rotate =#1, fill=gray!10]
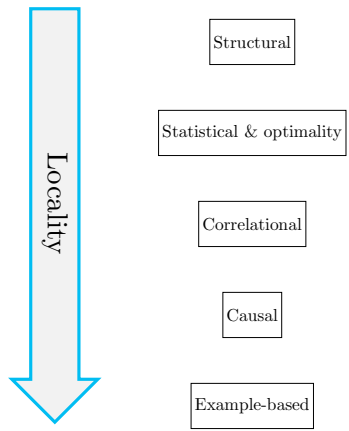
\begin{figure}[H]
\centering
\resizebox{2.5cm}{!}{\begin{tikzpicture}[every text node part/.style={align=center}]
    \node[main node, rectangle, inner sep=0.5ex] (1) {Structural};
    \node[main node, rectangle, inner sep=0.5ex] (2) [below = 1cm of 1] {Statistical \& optimality};
    \node[main node, rectangle, inner sep=0.5ex] (3) [below = 1cm of 2]  {Correlational};
    \node[main node, rectangle, inner sep=0.5ex] (4) [below = 1cm of 3] {Causal};
    \node[main node, rectangle, inner sep=0.5ex] (5) [below = 1cm of 4]  {Example-based};
\end{tikzpicture}}
\begin{tikzpicture}[overlay]
\node at (-4,3) [my arrow=-180] {\rotatebox{-90}{Locality}};
\end{tikzpicture}
\caption{Hierarchical levels of AI explanations.}
\end{figure}

Although in \emph{some} cases, the generation of minimal explanations is sufficient in satisfying the explanatory goals, a maximalist connception of explanation appears more truthworthy in the assessment of the social implications of AI decision-making. Moving from a maximalist to a minimalist account is simpler than branching out from a minimalist conception of explanations to several technical elements that determine the algorithmic decision outcomes. My argument thus does not imply that maximal explanations are required in \emph{all} cases of algorithmic decision-making. When the outcome has no significant impact, maximal explanations might be pointless. However, the framework emphasizes that there are multiple (empirically or mathematically) true reasons for why an algorithm arrives at a decision outcome. In critical decision-making contexts in which several politically and morally significant considerations become relevant, the maximalist conception of explanation is useful.

Recall that explanations are responses to why questions and are somehow, empirically or mathematically, true. If these explanations encompass statistical laws and rules, or mathematical optimization, we acquire mathematical and non-data driven explanations. These two kinds of explanations open an important yet neglected discourse about the legitimacy of the (partial) mathematical construction of decision outcomes in sensitive contexts. The proposed explanatory hierarchy enforces us to convey the explicit and implicit mathematical and causal assumptions that designers of the algorithms grapple with, as well as the objective functions in terms of which they design the algorithm.

In descending order of locality, there are three other types of explanation added to the hierarchy. These levels correspond to information obtained from the data-analysis (i.e., data-driven explanations) rather than an emphasis on the mathematical and statistical constitution of the decision-making process. These are correlational explanations, causal explanations, and example-based explanations. A hierarchy for data-driven explanations can be extracted from the discussion of Pearl concerning the hierarchy of causation.\footfullcite{pearl2000causality} Therefore, my proposed hierarchy (Figure 5) can be seen as an extension of Pearl's hierarchy of causation to different types of data-based and non-data based explanations for why and how an algorithmic decision outcome is obtained.

The explanatory schema recommends a pluralistic attitude about explanations: explanations complement each other, and shed light on different aspects of how a decision outcome is achieved. The complementary rather than the competitive nature of explanations distances us from discussions about whether one unique AI explanation is the correct one or not. This attitude is intimately connected to immediate issues about the social implications of AI decision-making. 

Are the receivers of explanations those who design the algorithms, the corporate head of a private company who has requested the algorithms to decide on an optimally accurate basis for some decision-making problems, the laypeople who do not have any knowledge of formal representations and mathematical tools? If so, what kinds of moral and political rules and values are implicitly acknowledged when an algorithm legitimately decides in sensitive contexts? Are liberal-democratic, socialistic, or corporate authoritarian rules and values acknowledged in the justification of these AI decision outcomes? The responses to such questions reveal the significance of the accepted background assumptions concerning the algorithmic decision-making. Hence, the explanatory hierarchy has the value of revealing the significance of explicit and implicit assumptions baked into AI decision-making. This suggests that each of the explanatory levels (illustrated in Figure 5) is indeed accompanied with some implicit and explicit background assumptions for their acceptability. If all the background assumptions required for algorithmic decision-making and the receiver of an explanation are matched, AI provides successful and effective explanations for an algorithmic outcome. 


So far, I have argued that there are several types of explanations conveying various grades of data-driven, non-data driven, and mathematical information about why and how a decision outcome is achieved. In the next section, I develop a conceptual framework that ties the varieties of explanations to the required background assumptions for the explanatory judgements. Often in the literature, the two concepts of machine-learning explainability and interpretability are defined in terms of each other. Here are two examples: ``In the context of ML systems, I define interpretability as the ability to explain or to present in understandable terms to a human.''\footfullcite{doshi2017towards} ``Explanation is thus one mode in which an observer may obtain understanding, but clearly, there are additional modes that one can adopt, such as making decisions that are inherently easier to understand or via introspection. I equate `interpretability' with `explainability'.''\footfullcite{miller2018explanation} As we will see, there are good reasons to distinguish these two concepts.

\section{Background assumptions and explanatory judgements}

The production of explanations about decisions made by AI systems is not the end of the AI explainability and interpretability debate. The practical value of these explanations, partly, depends on the audience who consumes them: an explanation must result in an appropriate level of understanding or some grade of cognitive achievement for the receivers of explanations. In other words, explanations are required to be interpreted and judged against different vantage points, about whether they are good or bad, satisfactory or unsatisfactory, effective or ineffective, acceptable or unacceptable. Among those background assumptions that impact the judgements of explanations are norms. It has been empirically shown that norms influence causal judgements.\footfullcite{bear2017normality} To put it simply, norms are informal rules that are held by people, and can have statistical or prescriptive content. Here is an example of how a norm can be relevant to the evaluation of an explanation, and yet can disagree with the explanation. 

Let us suppose that I have some strictly deontological commitments for assessing the moral permissibility of an action. In judging an explanation for a moral decision-making outcome based on an act-utilitarian solution, I have reasons to resist such an explanation due to disagreements with the style of reasoning for evaluating the moral permissibility of this act. Going back to Nora's example, suppose that the algorithm takes five measurable features of Nora into account, and finds correlations between the input features and the decision outcome. The correlations make up the explanation for why a decision outcome is achieved. Let us imagine an institution that has strong reasons to believe that Nora has a feature such as dignity, relevant to the decision-making context, and that dignity is incommensurable with the quantified input features required by the deep supervised learning for its decision-making. The institution then disagrees with the relevant explanation for why Nora is rejected due to the mismatch between its background assumptions (dignity as a non-quantifiable feature relevant to the decision-making context) and the algorithm assumption that the salient features to decision-making are all quantifiable and are input features to the algorithm.

More generally, the empirical and mathematical aspects for why a decision outcome is achieved are interpreted against some background assumptions held by the audiences of explanations. Some disagreements with an explanation for a decision outcome in a sensitive context due to the background assumptions of the audience of explanations reveal some moral or social mismatch about algorithmic decision-making between the receiver of an explanation and its producer. The fact that background assumptions impact explanatory judgements motivate why I tackle the interpretability of explanation as a separate, yet closely related issue to explanation. If one does not have a proper level of knowledge about the relevant precedent assumptions, one might not have the capacity to judge and interpret an explanation of a decision. The interpretability of explanations has a significant practical value for revealing the explicit and the implicit reasons about why a decision-making procedure and process is chosen. 



A schema for the interpretability of explanations aims to capture various precedent assumptions that become relevant in context-dependent evaluation of each kind of AI explanation for why a decision outcome is achieved. Back to the example of Nora, here are a list of moral and social considerations tied into technical aspects of algorithmic decision-making: Why does the algorithm merely observe Nora's quantified attributes relevant to decision-making? That is, what legitimizes the reductive-to-quantities representation of Nora's attributes, relevant to the hiring context? Why does an amalgamation of statistical and optimization-based reasoning decide about Nora's hire? Why is this learning algorithm used rather than one more favourable to Nora? If the algorithm is justified, why these rather than those (hyper-)parameters? Why does a training data set about others influence Nora's condition? That is, why does this training data set, rather than another (with having a different error distribution), impact Nora's condition?)

Associated with the structural explanation, there is an interpretative level which encompasses several assumptions for the acceptability of the reduction of the salient features of a decision-making problem to quantifiable measures. Statistical explanations are evaluated against drawing a distinction between statistical (and optimality) reckoning vs. non-statistical judgements, and the acceptability of statistical (and optimality) reckoning is tied to background assumptions for this legitimacy. Correlational explanations must be evaluated against various interventionist goals and norms: why there is a correlation, say, between Nora's ethnicity and why she is rejected for the employment position. Associated with causal explanations, there are some assumptions concerning the appropriateness of using causal discovery algorithms relevant to a kind of a decision-making problem. Example-based explanations are judged and evaluated against background assumptions such as the appropriateness and inclusiveness of the set of training data.

\tikzstyle{my arrow} = [draw=cyan!75, very thick, single arrow, minimum height=5.5cm, shape border rotate =#1, fill=gray!10]
\begin{figure}[H]
\centering
\resizebox{12cm}{!}{\begin{tikzpicture}[every text node part/.style={align=center}]

    \node[main node, rectangle, inner sep=0.5ex] (1) {Structural};
    \node[main node, rectangle, inner sep=0.5ex] (2) [below = 1cm of 1] {Statistical \& optimality};
    \node[main node, rectangle, inner sep=0.5ex] (3) [below = 1cm of 2]  {Correlational};
    \node[main node, rectangle, inner sep=0.5ex] (4) [below = 1cm of 3] {Causal};
    \node[main node, rectangle, inner sep=0.5ex] (5) [below = 1cm of 4]  {Example-based};
\node at (-4,-4) [my arrow=-180] {\rotatebox{-90}{Locality}};

    \node[main node, rectangle, inner sep=0.5ex] (6) [right =3.7cm of 1] {Mathematical reduction of decision problem};
    \node[main node, rectangle, inner sep=0.5ex] (7) [right =2cm of 2] {Statistical reckoning vs. non-statistical judgement};
    \node[main node, rectangle, inner sep=0.5ex] (8) [right =4.2cm of 3] {Norms and intervention goals};    
    \node[main node, rectangle, inner sep=0.5ex] (9) [right =5cm of 4] {Discovering causal relations};       
    \node[main node, rectangle, inner sep=0.5ex] (10) [right =4.5cm of 5] {Reference class problem};  
    \path[draw,loosely dashed,thick, right=8pt]
    (1) edge node {} (6);
    \path[draw,loosely dashed,thick, right=8pt]
    (2) edge node {} (7);
    \path[draw,loosely dashed,thick, right=8pt]
    (3) edge node {} (8);
    \path[draw,loosely dashed,thick, right=8pt]
    (4) edge node {} (9);
    \path[draw,loosely dashed,thick, right=8pt]
    (5) edge node {} (10);
\end{tikzpicture}}
\caption{A conceptual framework for AI explainability-interpretability}
\end{figure}
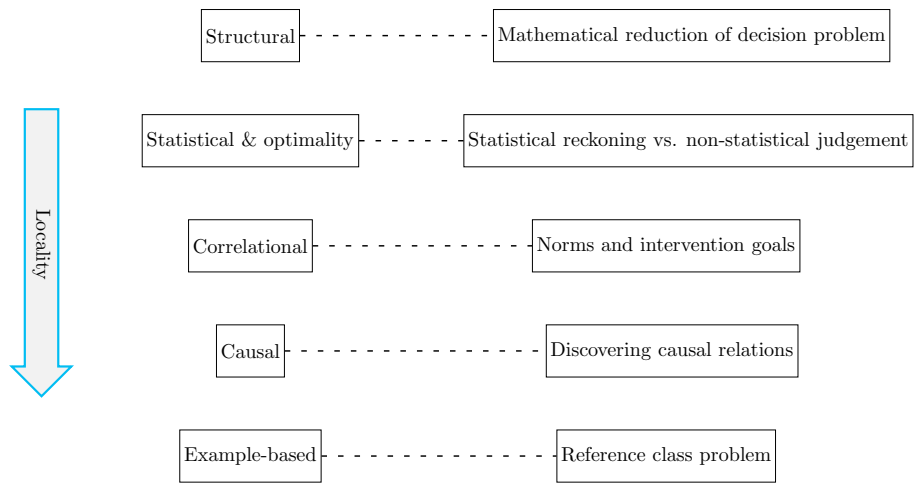

The hierarchical framework of explainability and interpretability, illustrated in Figure 6, suggests a comprehensive standard for assessment of the social, moral, and political aspects to which algorithmic decision-making pertains.


For instance, there is an increasing amount of discussion on the ethical and social implications of AI and algorithmic decision-making. One principled way to bring focus to this discussion is to benefit from the significance of the types of AI explanations for why an outcome is achieved. Mapping the AI explanations on the implicitly endorsed assumptions in critical and sensitive algorithmic decision-making contexts triggers the discussion on whether the algorithmic decision-making for a specific social domain is aligned with some endorsed moral values or not. I have used Nora's hiring problem as a running example in order to convey the importance of each element of AI explanations. The usefulness of the framework illustrated in Figure 6 applies more generally, to algorithmic decision-making in other sensitive domains such as criminal recidivism and medical diagnosis.


First, the most global level of the explanatory hierarchy corresponds to the mathematical and quantifiable definition of the relevant elements of the decision problem. As discussed in Section 2.1, deep supervised learning algorithms employ artificial neural networks with multiple hidden layers to represent the relevant aspects of the decision task. Whether or not the combination of nodes, arcs, their associated weights, the activation functions, and the past historical data can fully capture the relevant elements engaged in a particular decision-making problem is a legitimate question that requires a careful investigation. 



Second, a statistical explanation sheds light on the average behavior of a population sample. Hence, by definition statistical information and inferences are about group behaviors. For instance, the law of large numbers and the central limit theorem hold true only for a large data corpus, rather than a single individual. Hence, the decision outcomes that are obtained from statistical facts suffer from limitations concerning the explanation of why a decision, at an individualistic level, is obtainable. More specifically, there is a hard question about why a highly probabilistic outcome justifies the decision result at an individualistic level. I do not suggest that statistical limitations are always problematic. We use such inferences all the time, and indeed the contemporary scientific inquiry is based on this mode of reasoning. The correspondence between the second level of the explanatory hierarchy and its associated background assumptions presses the importance of highlighting reasons about why to adopt statistical reckoning vs. non-statistical judgement in context-sensitive decision-making problems.

Third, the correlational explanations suggest information about relevant features in the decision-making problem, but these explanations might not be fully illuminating, given that the choices of mathematical elements determine why a decision outcome is achieved. They do not provide a deeply convincing argument for why a decision should be made; rather, they make inferences about a particular case, based on how some random variables in a training data set are correlated. The correlational explanations also pave the way for discovering causal information from a set of data. 

Fourth, in the context of machine learning, causal information is extracted, according to some strong assumptions about our prior probabilistic knowledge. The information about probabilistic causality are acquired from correlational information embedded in the observational data. The causal information that is extracted, by methods such as do-calculus, is model-dependent.\footfullcite{pearl2018book} They mainly have significant value for interventions.

Fifth, example-based explanations reveal some information about a decision context. They might manifest counterfactual information, or just contrastive information. 

An example-based explanation for why my credit card application is rejected might be that if I have had the age and salary of John, my application would have been approved. These explanations might be understandable by lay people, but they do not deeply show why a decision outcome based on some rules take place. Keeping the level of discussion at searching for a minimal information that is understandable by lay people might allow private sectors to hide more important information that is the true main reason for why a decision outcome is achieved. 

\section{Conclusion}

Due to the fact that algorithmic decision-making has the potential to affect detrimentally individual or collective human rights and situations, the AI explanations deserve more significant analysis than those systems, using decision-making algorithms to process simply objects. In this paper, I proposed a hierarchical framework for the explainability and interpretability of AI, by enumerating a variety of explanations. The proposed framework provides a platform to bring some systematic focus to the many kinds of questions we might ask about deep decision-making in sensitive contexts. Some argue that the interpretability and explainability of AI has been motivated by generating post-hoc explanations that are understandable by lay people. I resist this narrow conception of explainability and interpretability for sensitive decision contexts. To be clear, I do not claim that in all decision contexts, the aforementioned levels of explanation and their corresponding elements of interpretation are relevant. For instance, on classifying the images of a cat, the relevant explanation-interpretation discussion can significantly be shortened. On the other hand, using a deep supervised learning algorithm for hiring employees or criminal recidivism demands a more elaborate clarification of the aspects of algorithmic decision-making. The conceptual framework illustrated in Figure 6 has enough resources to capture these elements. Moreover, paying attention to the background normative assumptions provides insights for computer scientists into how to operationalize AI explanations.


@inproceedings{mittelstadt2019explaining,
  title={Explaining explanations in AI},
  author={Mittelstadt, Brent and Russell, Chris and Wachter, Sandra},
  booktitle={Proceedings of the conference on fairness, accountability, and transparency},
  pages={279--288},
  year={2019},
  organization={ACM}
}

@article{cabitza2017unintended,
  title={Unintended consequences of machine learning in medicine},
  author={Cabitza, Federico and Rasoini, Raffaele and Gensini, Gian Franco},
  journal={Jama},
  volume={318},
  number={6},
  pages={517--518},
  year={2017},
  publisher={American Medical Association}
}

@book{pearl2000causality,
  title={Causality: models, reasoning and inference},
  author={Pearl, Judea},
  volume={29},
  year={2000},
  publisher={Springer}
}

@article{coella2019,
 author  = {Moran, Lee},
 date    = {2019-08-06},
 title   = {Judge Tosses Speeding Ticket Of 96-Year-Old Man Caring For Son With Cancer.},
 journal = {Hoffington Post}
 }

@article{batterman2014minimal,
  title={Minimal model explanations},
  author={Batterman, Robert W and Rice, Collin C},
  journal={Philosophy of Science},
  volume={81},
  number={3},
  pages={349--376},
  year={2014},
  publisher={University of Chicago Press}
}

@article{potochnik2007optimality,
  title={Optimality modeling and explanatory generality},
  author={Potochnik, Angela},
  journal={Philosophy of Science},
  volume={74},
  number={5},
  pages={680--691},
  year={2007},
  publisher={The University of Chicago Press}
}

@article{bokulich2011scientific,
  title={How scientific models can explain},
  author={Bokulich, Alisa},
  journal={Synthese},
  volume={180},
  number={1},
  pages={33--45},
  year={2011},
  publisher={Springer}
}

@article{zhao2019causal,
  title={Causal interpretations of black-box models},
  author={Zhao, Qingyuan and Hastie, Trevor},
  journal={Journal of Business \& Economic Statistics},
  pages={1--19},
  year={2019},
  publisher={Taylor \& Francis}
}

@inproceedings{lakkaraju2017learning,
  title={Learning cost-effective and interpretable treatment regimes},
  author={Lakkaraju, Himabindu and Rudin, Cynthia},
  booktitle={Artificial Intelligence and Statistics},
  pages={166--175},
  year={2017}
}

@inproceedings{ribeiro2016should,
  title={Why should i trust you?: Explaining the predictions of any classifier},
  author={Ribeiro, Marco Tulio and Singh, Sameer and Guestrin, Carlos},
  booktitle={Proceedings of the 22nd ACM SIGKDD international conference on knowledge discovery and data mining},
  pages={1135--1144},
  year={2016},
  organization={ACM}
}

@incollection{smith1994law,
  title={The law of large numbers and the strength of insurance},
  author={Smith, Michael L and Kane, Stephen A},
  booktitle={Insurance, risk management, and public policy},
  pages={1--27},
  year={1994},
  publisher={Springer}
}

@inproceedings{kim2016examples,
  title={Examples are not enough, learn to criticize! criticism for interpretability},
  author={Kim, Been and Khanna, Rajiv and Koyejo, Oluwasanmi O},
  booktitle={Advances in Neural Information Processing Systems},
  pages={2280--2288},
  year={2016}
}

@book{pearl2018book,
  title={The book of why: the new science of cause and effect},
  author={Pearl, Judea and Mackenzie, Dana},
  year={2018},
  publisher={Basic Books}
}

@article{chirimuuta2017explanation,
  title={Explanation in computational neuroscience: Causal and non-causal},
  author={Chirimuuta, Mazviita},
  journal={The British Journal for the Philosophy of Science},
  volume={69},
  number={3},
  pages={849--880},
  year={2017},
  publisher={Oxford University Press}
}

@article{bear2017normality,
  title={Normality: Part descriptive, Part prescriptive},
  author={Bear, Adam and Knobe, Joshua},
  journal={Cognition},
  volume={167},
  pages={25--37},
  year={2017},
  publisher={Elsevier}
}

@book{reutlinger2018explanation,
  title={Explanation beyond causation: philosophical perspectives on non-causal explanations},
  author={Reutlinger, Alexander and Saatsi, Juha},
  year={2018},
  publisher={Oxford University Press}
}

@article{rumelhart1988learning,
  title={Learning representations by back-propagating errors},
  author={Rumelhart, David E and Hinton, Geoffrey E and Williams, Ronald J and others},
  journal={Cognitive modeling},
  volume={5},
  number={3},
  pages={1},
  year={1988}
}

@misc{angwin2016machine,
  title={Machine Bias: there’s software used across the country to predict future criminals. and it’s biased against blacks. ProPublica 2016},
  author={Angwin, Julia and Larson, Jeff and Mattu, Surya and Kirchner, Lauren},
  year={2016}
}

@article{hempel1948studies,
  title={Studies in the Logic of Explanation},
  author={Hempel, Carl G and Oppenheim, Paul},
  journal={Philosophy of Science},
  volume={15},
  number={2},
  pages={135--175},
  year={1948},
  publisher={Williams and Wilkins Co.}
}

@article{hempel1965aspects,
  title={Aspects of Scientific Explanation; And Other Essays in the Philosophy of Science},
  author={Hempel, Carl G},
  year={1965}
}

@article{rice2015moving,
  title={Moving beyond causes: Optimality models and scientific explanation},
  author={Rice, Collin},
  journal={No{\^u}s},
  volume={49},
  number={3},
  pages={589--615},
  year={2015},
  publisher={Wiley Online Library}
}

@book{lange2016because,
  title={Because without cause: Non-causal explanations in science and mathematics},
  author={Lange, Marc},
  year={2016},
  publisher={Oxford University Press}
}

@book{Nielsen2015,
  title={Neural networks and deep learning},
  author={Nielsen, Michael A},
  year={2015},
  publisher={Determination Press}
}

@book{batterman2001devil,
  title={The devil in the details: Asymptotic reasoning in explanation, reduction, and emergence},
  author={Batterman, Robert W},
  year={2001},
  publisher={Oxford University Press}
}

@book{salmon1984scientific,
  title={Scientific explanation and the causal structure of the world},
  author={Salmon, Wesley C},
  year={1984},
  publisher={Princeton University Press}
}

@book{strevens2008depth,
  title={Depth: An account of scientific explanation},
  author={Strevens, Michael},
  year={2008},
  publisher={Harvard University Press}
}

@article{bodo2017tackling,
  title={Tackling the algorithmic control crisis-the technical, legal, and ethical challenges of research into algorithmic agents},
  author={Bod{\'o}, Bal{\'a}zs and Helberger, Natalie and Irion, Kristina and Zuiderveen Borgesius, F and Moller, Judith and van de Velde, Bob and Bol, Nadine and van Es, Bram and de Vreese, Claes},
  journal={Yale Journal of Law \& Technology},
  volume={19},
  pages={133},
  year={2017},
  publisher={HeinOnline}
}

@article{veale2018clarity,
  title={Clarity, surprises, and further questions in the Article 29 Working Party draft guidance on automated decision-making and profiling},
  author={Veale, Michael and Edwards, Lilian},
  journal={Computer Law \& Security Review},
  volume={34},
  number={2},
  pages={398--404},
  year={2018},
  publisher={Elsevier}
}

@book{Cantwellsmith,
  title={The promise of Artificial Intelligence},
  author={Smith, Brian Cantwell},
  year={2019},
  Publisher={MIT University Press}
}

@article{lecun2015deep,
  title={Deep learning},
  author={LeCun, Yann and Bengio, Yoshua and Hinton, Geoffrey},
  journal={Nature},
  volume={521},
  number={7553},
  pages={436},
  year={2015},
  publisher={Nature Publishing Group}
}

@article{doshi2017towards,
  title={Towards a rigorous science of interpretable machine learning},
  author={Doshi-Velez, Finale and Kim, Been},
  journal={arXiv preprint arXiv:1702.08608},
  year={2017}
}

@article{lipton2016mythos,
  title={The mythos of model interpretability},
  author={Lipton, Zachary C},
  journal={arXiv preprint arXiv:1606.03490},
  year={2016}
}

@article{wachter2017counterfactual,
  title={Counterfactual Explanations without Opening the Black Box: Automated Decisions and the GPDR},
  author={Wachter, Sandra and Mittelstadt, Brent and Russell, Chris},
  journal={Harvard Journal of Law \& Technology},
  volume={31},
  pages={841},
  year={2017},
  publisher={HeinOnline}
}

@article{marcus2018deep,
  title={Deep learning: A critical appraisal},
  author={Marcus, Gary},
  journal={arXiv preprint arXiv:1801.00631},
  year={2018}
}

@article{miller2018explanation,
  title={Explanation in artificial intelligence: Insights from the social sciences},
  author={Miller, Tim},
  journal={Artificial Intelligence},
volume={267},
  pages={1--38},
  year={2018},
  publisher={Elsevier}
}

@incollection{Bromberger1966,
  title={Why-Questions},
  author={Bromberger, Sylvain},
  booktitle={Mind and cosmos: Essays in contemporary science and philosophy},
  editor={Colodny, Robert G},
  pages={86--111},
  year={1966}
}
\end{document}